\documentclass[runningheads]{llncs}

\usepackage{graphicx}
\usepackage{comment}
\usepackage{amsmath,amssymb}
\usepackage{color}
\usepackage{url}
\usepackage{hyperref}

\usepackage{enumitem}
\usepackage{multirow}
\usepackage{amsmath}
\usepackage{diagbox}
\usepackage{placeins}
\usepackage{breakcites}

\newif\ifreview
\reviewfalse

\ifreview
	\usepackage{lineno}

	\linenumbers
\fi

\begin{document}

%%%%%%%%%%%%%%%%%%%%% Add submission id, track, and title. %%%%%%%%%%%%%%%%%%%%%

% TODO: Please insert your submission number here
\def\SubNumber{44}

% TODO: Please uncomment the track this paper will be submitted to, comment all other lines
\def\GCPRTrack{Main Track}
%\def\GCPRTrack{Special Track: Computer vision systems and applications}
%\def\GCPRTrack{Special Track: Pattern recognition in the life and natural sciences}
%\def\GCPRTrack{Special Track: Photogrammetry and remote sensing}
%\def\GCPRTrack{Special Track: Robot vision}
%\def\GCPRTrack{Young Researcher's Forum}
%\def\GCPRTrack{Fast Review Track}

% TODO: Replace with your title
\title{ArtFID: Quantitative Evaluation of Neural Style Transfer}
% You can use \thanks for acknowledgment. Do not add any acknowledgment to the draft 
% version that is used for the review process.  
%\title{Title\thanks{XXX}}

\ifreview
	% ANONYMOUS SUBMISSION FOR REVIEW
	% DO NOT MODIFY these for the draft version that is used for the review process.
	\titlerunning{GCPR 2022 Submission \SubNumber{}. CONFIDENTIAL REVIEW COPY.}
	\authorrunning{GCPR 2022 Submission \SubNumber{}. CONFIDENTIAL REVIEW COPY.}
	\author{GCPR 2022 - \GCPRTrack{}}
	\institute{Paper ID \SubNumber}
\else

	\author{Matthias Wright\inst{1}\and
	Bj\"{o}rn Ommer \inst{1,2}}
	
	\authorrunning{M. Wright et al.}
	
	\institute{Ludwig Maximilian University of Munich, Germany \and IWR, Ruprecht Karl University of Heidelberg, Germany\\
	\email{m.wright@campus.lmu.de}\\
	\url{https://github.com/matthias-wright/art-fid}}
\fi

\maketitle              % typeset the header of the contribution

\begin{abstract}
The field of neural style transfer has experienced a surge of research exploring different avenues ranging from optimization-based approaches and feed-forward models to meta-learning methods. The developed techniques have not just progressed the field of style transfer, but also led to breakthroughs in other areas of computer vision, such as all of visual synthesis. However, whereas quantitative evaluation and benchmarking have become pillars of computer vision research, the reproducible, quantitative assessment of style transfer models is still lacking. Even in comparison to other fields of visual synthesis, where widely used metrics exist, the quantitative evaluation of style transfer is still lagging behind. To support the automatic comparison of different style transfer approaches and to study their respective strengths and weaknesses, the field would greatly benefit from a quantitative measurement of stylization performance. Therefore, we propose a method to complement the currently mostly qualitative evaluation schemes. We provide extensive evaluations and a large-scale user study to show that the proposed metric strongly coincides with human judgment.

\end{abstract}

\section{Introduction}
\begin{figure}
  \centering
  \includegraphics[width=0.9\textwidth]{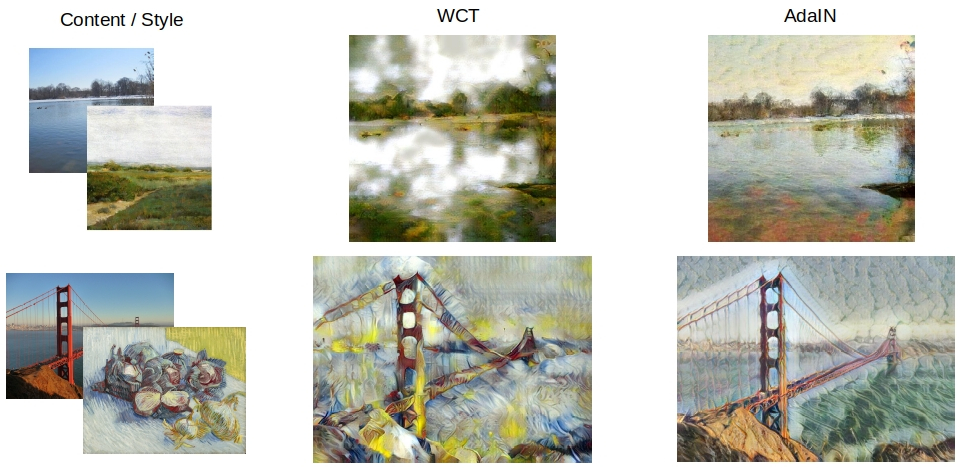}
  \caption{AdaIN \cite{Huang2017ArbitraryST} (arguably) performs better on the top style/content pair, while WCT \cite{Li2017UniversalST} (arguably) performs better on the bottom style/content pair.}
  \label{fig:wct_adain}
\end{figure}

Style transfer and texture transfer have been researched at least since the early 2000s \cite{Hertzmann2001,Efros2001}. In 2016, Gatys et al. proposed a novel method for style transfer based on the features of a pretrained convolutional neural network. The recent years have seen a surge of research exploring different avenues including, but not limited to, optimization-based approaches \cite{Gatys2016,Kolkin2019,Kim2020DST,Kotovenko2021}, feed-forward models \cite{Johnson2016Perceptual,Ulyanov2016TextureNF}, universal feed-forward models \cite{Huang2017ArbitraryST,Li2017UniversalST,Chen2021ST,liu2021adaattn,Park2019ST,huo2021manifold,sheng2018avatar,li2018learning}, ultra-resolution techniques \cite{Wang2020CollaborativeDF,chen2022towards}, meta-learning approaches \cite{Shen2018meta,zhang2019metastyle}, and video style transfer \cite{ruder2018,Chen2020OpticalFD,Huang2017video,Chen2017video}.\\
The developed techniques have not just progressed the field of style transfer, but also led to breakthroughs in other areas of computer vision, such as visual synthesis. For example, the perceptual loss \cite{Gatys2016,Johnson2016Perceptual} has been applied to tasks ranging from photographic image synthesis \cite{Chen2017Photographic} to motion transfer \cite{Chan2019Dance}, the generator architecture employed by Johnson et al. \cite{Johnson2016Perceptual} has been used to synthesize photo-realistic images from semantic label maps \cite{Wang2018Semantic} or for unpaired image-to-image translation \cite{Zhu2017Cycle}, and the AdaIN layer \cite{Huang2017ArbitraryST} was a key ingredient for the StyleGAN generator \cite{stylegan2019} that produced state-of-the-art results in unconditional generative image modeling. \\
In contrast to other areas of computer vision research such as classification or even natural image synthesis, where quantitative evaluation and benchmarking are already common practice, quantitative assessment of style transfer models is still lacking.\\
Currently, the most common method for evaluating these models is a qualitative comparison for a few, commonly used, style and content images. This is generally useful for an initial impression, however, the performance of style transfer models greatly varies across different style and content images. For example, Fig.~\ref{fig:wct_adain} shows stylized images from AdaIN \cite{Huang2017ArbitraryST} and WCT \cite{Li2017UniversalST} for two different style/content pairs. One can argue that AdaIN \cite{Huang2017ArbitraryST} outperforms WCT \cite{Li2017UniversalST} for the first pair and vice versa for the second pair.\\
Some papers provide a quantitative comparison using measurements for speed, memory, or control \cite{chen2022towards,Huang2017ArbitraryST,Shen2018meta,Luo2022ConsistentST,Li2017UniversalST,Ulyanov2016TextureNF,zhang2019metastyle,Hong_2021_ICCV,Chiu2020IterativeFT,liu2021paint,Wu_2021_ICCV,jcyang2022InST,Liu2021warp,Chiu2022}. These measurements are very helpful to highlight a model's performance with respect to a specific property such as speed, but are generally not intended to measure stylization performance. A few works \cite{Hong_2021_ICCV,Kim2022prior,Liu_Wu_Wu_Wen_2021,jcyang2022InST} also employ classical perceptual metrics such as PSNR or SSIM \cite{Wang2004ssim}, however, these metrics are generally not consistent with human perception \cite{zhang2018perceptual}.\\
Sanakoyeu et al. \cite{sanakoyeu2018styleaware} proposed the \textit{Deception rate} to measure the quality of stylized images. It is the fraction of stylized images that an artist classification network has assigned to the artist, whose artwork has been used for stylization. The deception rate can be useful in some cases but also suffers from certain drawbacks and is not widely used as a result. First, it only works for style images belonging to an artist for which the artist classifier network has been trained and thus cannot be used for arbitrary styles. And second, it completely disregards content preservation.\\
In the absence of a suitable quantitative measurement of stylization performance, many authors have turned to human evaluation studies \cite{Kolkin2019,Luo2022ConsistentST,Hong_2021_ICCV,Chandran2021ada,Park2019ST,sanakoyeu2018styleaware,liu2021adaattn,Kotovenko2021,Wang2020CollaborativeDF,Kim2022prior,Chen_2021_ICCV,Xu2021DRBGANAD,Wu2020align,Chen2021ST,jcyang2022InST,deng2020multiadapt,yang2022Pastiche,deng2022stytr2,Kwon2022}, where users are asked to compare stylized images of different models and specify their preference. These studies give better insights into a model's stylization performance in comparison to other methods, especially when conducted with many users and for many different style and content images. Nonetheless, there are cases where a user study is not feasible. Large-scale crowdsourcing can be expensive and may not be an option for researchers with limited funding. Furthermore, there are cases where a quantitative measurement is simply more suitable. As an example, consider the closely-related field of image synthesis, where quantitative measurements for image quality are used, not just for evaluation, but also as a core part of the approach. Recent examples include work by Mokady et al. \cite{mokady2022selfdistilled}, where Fr\'{e}chet Inception Distance (FID) \cite{Heusel2017} and Learned Perceptual Image Patch Similarity (LPIPS) \cite{zhang2018perceptual} measurements are employed for self-filtering a collection of images, work by Karras et al. \cite{Karras2020ada} that use the FID to detect GAN \cite{goodfellow2014gan} overfitting, or a large-scale GAN study by Lucic et al. \cite{lucic2018are} that utilize the FID to compare the sensitivity of different GAN models to hyper-parameters. A quantitative measurement of stylization performance would facilitate similar studies to further analyze and improve style transfer methods. In this work, we propose a method that fills this role and call it \textit{ArtFID}. Our goal is not to replace the current evaluation techniques, but rather to complement them.\\
Stylization performance is mostly determined by two factors: \textit{content preservation} and \textit{style matching}. Content preservation refers to the extent to which the semantic content from the content images is preserved in the stylized images. Style matching refers to the extent to which the style of the generated images resembles the style of the target style images. Measuring content preservation is straightforward using the well-established perceptual loss \cite{Gatys2016,Johnson2016Perceptual}, CLIP loss \cite{clip2021}, or LPIPS metric \cite{zhang2018perceptual}. However, measuring style matching is more challenging because it confronts us with the question "what is style?", a question that has been debated relentlessly by art historians \cite{kubler1979,ackerman1962,wallach1997,meyer1979,schapiro1953style}.
However, a commonly used definition is that style is the "distinctive manner which permits the grouping of works into related categories" \cite{Fernie1995}. This definition gives rise to a representation learning approach, where artworks are classified into categories by a neural network to obtain representations that reflect their stylistic similarity. To this end, we collect a large-scale dataset of artworks, labeled by artist and stylistic period. We make use of both labels by employing an architecture with two classification heads. Having obtained the image representations in the form of neural network features, we return to the issue of measuring style matching. As shown by Li et al. \cite{li2017demystifying}, the task of style transfer corresponds to aligning feature distributions of neural networks. To measure style matching, we can thus measure how well the feature distribution of style images matches the feature distribution of stylized images. Following the work by Heusel et al. \cite{Heusel2017}, we measure the distance between these two feature distributions with respect to the first two moments using the Fr\'{e}chet distance \cite{Frechet1957,DOWSON1982450}. \\
Our contributions are as follows:
\begin{itemize}[itemsep=0.2em, topsep=0pt,parsep=0pt,partopsep=0pt, leftmargin=1.5em]
    \item We propose a method for quantitatively evaluating style transfer models with respect to stylization performance that strongly coincides with human judgment.
    \item We introduce a large-scale dataset, containing 250k images of labeled artworks.
    \item We provide extensive evaluations and a large-scale human evaluation study to support our claims.
\end{itemize}

\section{Related Work}

\paragraph{Style Transfer}
Research in the fields of style transfer and texture transfer goes back to the early 2000s. Efros and Freeman \cite{Efros2001} proposed texture synthesis and transfer using image quilting and Hertzmann et al. \cite{Hertzmann2001} proposed \textit{image analogies} based on a simple multi-scale autoregression. Other work in this area includes a method for synthesizing directional textures \cite{wang2004}. \\
In 2016, Gatys et al. \cite{Gatys2016} proposed an iterative optimization algorithm for style transfer that leverages the representations learned by convolutional neural networks. In the following, Johnson et al. \cite{Johnson2016Perceptual} and Ulyanov et al. \cite{Ulyanov2016TextureNF} used feed-forward networks to approximate the optimization problem formulated by Gatys et al. \cite{Gatys2016} for a specific style image.\\
In order to use mutliple styles with a single network, Dumoulin et al. \cite{Dumoulin2017} proposed conditional instance normalization. Huang and Belongie \cite{Huang2017ArbitraryST} proposed the first method for arbitrary style transfer, using adaptive instance normalization. Li et al. \cite{Li2017UniversalST} introduced another arbitrary style transfer model, based on whitening and colouring transformations. \\
Li et al. \cite{li2017demystifying} provided insights into the mechanics of style transfer by proving that matching the Gram matrices of feature maps is equivalent to minimizing the Maximum Mean Discrepancy with the second order polynomial kernel. \\
Li et al. \cite{Li2018ClosedForm} proposed a closed-form solution for photorealistic stylization that is based on the aforementioned whitening and colouring transformations. Sheng et al. \cite{sheng2018avatar} introduced a method that semantically aligns content features to trade off generalization and efficiency and  Li et al. \cite{li2018learning} proposed a linear transformation for image and video style transfer. \\
Shen et al. \cite{Shen2018meta} and Zhang et al. \cite{zhang2019metastyle} employed meta-learning approaches to handle the trade-off between speed, flexibility, and quality, and Svoboda et al. \cite{Svoboda2020cvpr} used graph neural networks to recombine style and content in latent space.
Another line of research used attention mechanisms to integrate local style patterns and align content and style manifolds \cite{Park2019ST,deng2020multiadapt,liu2021adaattn,Luo2022ConsistentST,huo2021manifold}. \\
Recent work includes exemplar-based portrait style transfer \cite{yang2022Pastiche}, online motion style transfer \cite{tao2022style}, transformer-based style transfer \cite{deng2022stytr2}, lightweight photorealistic style transfer \cite{Chiu2022}, style transfer for scene reconstruction \cite{hoellein2022stylemesh}, industrial style transfer \cite{jcyang2022InST}, thumbnail instance normalization for ultra-resolution style transfer \cite{chen2022towards}, style transfer based on text descriptions \cite{Kwon2022}, internal-external style transfer with two contrastive losses \cite{Chen2021ST}, reversible neural flows for unbiased stylization \cite{artflow2021}, and a unified architecture for domain-aware style transfer \cite{Hong_2021_ICCV}.

\paragraph{Inception Score}
Salimans et al. \cite{Salimans2016} proposed the \textit{Inception score}, an evaluation metric for generated images that correlates with human judgment. To compute the Inception score, the Inception network \cite{Szegedy2016} is used to compare the conditional label distribution to the marginal distribution of the generated images. A downside of the Inception score is that the statistics of the generated images are not compared to the statistics of real images \cite{Heusel2017}.

\paragraph{Fr\'{e}chet Inception Distance}
Heusel et al. \cite{Heusel2017} proposed the \textit{Fr\'{e}chet Inception Distance (FID)} that improves the Inception score by computing the Fr\'{e}chet distance \cite{Frechet1957,DOWSON1982450} between the Gaussian distribution of Inception features of generated images and the Gaussian distribution of Inception features of real images.

\section{Approach}\label{sec:approach}

\subsection{ArtFID}
As stated before, the stylization quality is determined by content preservation and style matching. In contrast to previous work, we measure both factors and combine them to form a quantitative metric.\\
Let $X_c$ denote the set of content images, $X_s$ the set of style images, and $X_g$ the set of stylized images, generated from a particular style transfer model. To measure content preservation for a particular stylized image, we compute a distance $d(X_c^{(i)}, X_g^{(i)})$ between the content image $X_c^{(i)}$ and the corresponding stylized image $X_g^{(i)}$. Possible choices for $d(\cdot)$ are the VGG perceptual loss \cite{Gatys2016,Johnson2016Perceptual}, the CLIP loss \cite{clip2021}, or the LPIPS metric \cite{zhang2018perceptual}. We select LPIPS and show in Sec.~\ref{sec:ablation} that it slightly outperforms the alternatives. \\
Measuring the content preservation for a style transfer model then simply corresponds to evaluating $d(X_c^{(i)}, X_g^{(i)})$ for a large batch of stylized images and taking the mean.\\
As stated above, to measure style matching for a model, we first learn suitable image representations by training a classifier on a large-scale artwork dataset (Sec.~\ref{sec:data}). For the classifier, we use the Inception network \cite{Szegedy2016}. The architecture is modified by using two classification heads instead of one. In doing so, we are able to harness the information from both the artist labels as well as the labels for the stylistic period that are present in the dataset. For more information about the classifier training, see Sec.~\ref{sec:training}. The image representations are the Inception features from the last layer before the classifier heads. These features are computed for both the style images $X_s$ and the stylized images $X_g$. To measure style matching between the style images $X_s$ and the stylized images $X_g$, we measure the distance between these two feature distributions. This is motivated by work from Li et al. \cite{li2017demystifying} that shows that the task of style transfer corresponds to aligning feature distributions of deep neural networks. Similar to Heusel et al. \cite{Heusel2017}, we measure the distance between these two feature distributions with respect to the first two moments using the Fr\'{e}chet distance \cite{Frechet1957,DOWSON1982450}
\begin{equation}\label{eq:fid}
    \textrm{FID}(X_s, X_g) = ||\mu_s - \mu_g||_2^2 + \textrm{Tr}(\Sigma_s + \Sigma_g - 2 (\Sigma_s \Sigma_g)^\frac{1}{2}),
\end{equation}
where $(\mu_s, \Sigma_s)$ and $(\mu_g, \Sigma_g)$ are the mean and covariance of the Inception features of the style images $X_s$ and the stylized images $X_g$, respectively. \\
Having obtained a measure for content preservation and a measure for style matching, we combine these measures to form the ArtFID, formulated as follows
\begin{equation}\label{eq:art_fid}
\textrm{ArtFID}(X_g, X_c, X_s) = \Bigg( 1 + \frac{1}{N} \sum\limits_{i=1}^N d(X_c^{(i)}, X_g^{(i)}) \Bigg) \cdot \Bigg( 1 + \textrm{FID}(X_s, X_g) \Bigg)
\end{equation}
Ones are added to both factors to avoid a degenerate minimum when $X_c = X_g$ or $X_s = X_g$. The measures are combined to ensure that only models that perform well on both content preservation and style matching achieve a low ArtFID score. Multiplication lends itself well to this purpose. We also tested the addition operation but the effect on the overall ranking was negligible.

\subsection{Unbiased ArtFID Computation}
Chong and Forsyth \cite{unbiasedfid2020} showed that the FID computed with a finite number of samples $N$, denoted FID$_N$, is biased. This bias in FID$_N$ vanishes when $N \rightarrow \infty$. The authors \cite{unbiasedfid2020} propose a method for computing $\overline{\textrm{FID}}_\infty$, an unbiased estimate of $\textrm{FID}_\infty$, using $N$ samples. The FID is computed $K$ times, each time with a different number of samples $M_1,...,M_K$. The $M_1,...,M_K$ are evenly spaced numbers over the interval $[5000, N]$. A linear regression model is then fitted to the points $(1 / M_1, \textrm{FID}_{M_1}),...,(1 / M_K, \textrm{FID}_{M_K})$. To obtain $\overline{\textrm{FID}}_\infty$, the linear regression model is evaluated at $0$. For an in-depth analysis, we refer to the work of Chong and Forsyth \cite{unbiasedfid2020}. \\
We use $\overline{\textrm{FID}}_\infty$ for computing the ArtFID and denote it as ArtFID$_\infty$. Tab.~\ref{tab:rank_correlation} shows that it performs slightly better than the biased ArtFID.

\subsection{Data}\label{sec:data}
The dataset used for training the Inception network consists of 250K images, labeled by artist and stylistic period. The images were collected from public datasets (e.g. WikiArt \cite{wikiart2016}), as well as museum databases and art collections (e.g. Art UK \footnote{\url{https://artuk.org/}}). See Sec.~\ref{app:dataset} for more details about the dataset.

\subsection{Training}\label{sec:training}
We broadly follow the methodology described by Szegedy et al. \cite{Szegedy2016} for training the Inception network. The classifier head is replaced by two separate heads, one for each label (artist and stylistic period). We use the Adam optimizer \cite{adam2015} with a learning rate of $0.0001$ and $\beta = (0.9, 0.999)$. The batch size is set to $64$ and the weight of the auxiliary classifier is set to $0.3$. See Sec.~\ref{app:training} for more information about the training and evaluation.

\section{Experiments}

\subsection{ArtFID Evaluation}
We evaluate the ArtFID with 13 different style transfer methods from the literature \cite{liu2021adaattn,sheng2018avatar,Chen2021ST,li2018learning,deng2020multiadapt,huo2021manifold,Luo2022ConsistentST,Huang2017ArbitraryST,Gatys2016,Park2019ST,Svoboda2020cvpr,chen2022towards,Li2017UniversalST}, some of which are well-established, some are recently published, and some are in between. For each method, we generate stylized images using the same pairs of content and style images. Content images are sampled from the Places365 dataset \cite{zhou2017places} and the COCO dataset \cite{mscoco2014}. Style images are sampled from the WikiArt dataset \cite{wikiart2016} and the BAM dataset \cite{Wilber_2017_ICCV}. Images are sampled from multiple datasets to increase the variability of style and content images. Following the recommendations by Heusel et al. \cite{Heusel2017}, we compute the ArtFID with samples of 50k images. Both style and content images are resized to 512x512, a commonly used image size for style transfer methods. As proposed by Parmar et al. \cite{parmar2022cleanfid}, we resize style and content images with bicubic interpolation and antialiasing and save the stylized images using the PNG format. \\
For each style transfer method, the ArtFID is computed with 5 different samples containing 50k images each. The results are reported in Tab.~\ref{tab:results}.

\subsection{Large-Scale Human Evaluation}\label{sec:amt}
\begin{figure}
  \centering
  \includegraphics[width=0.7\textwidth]{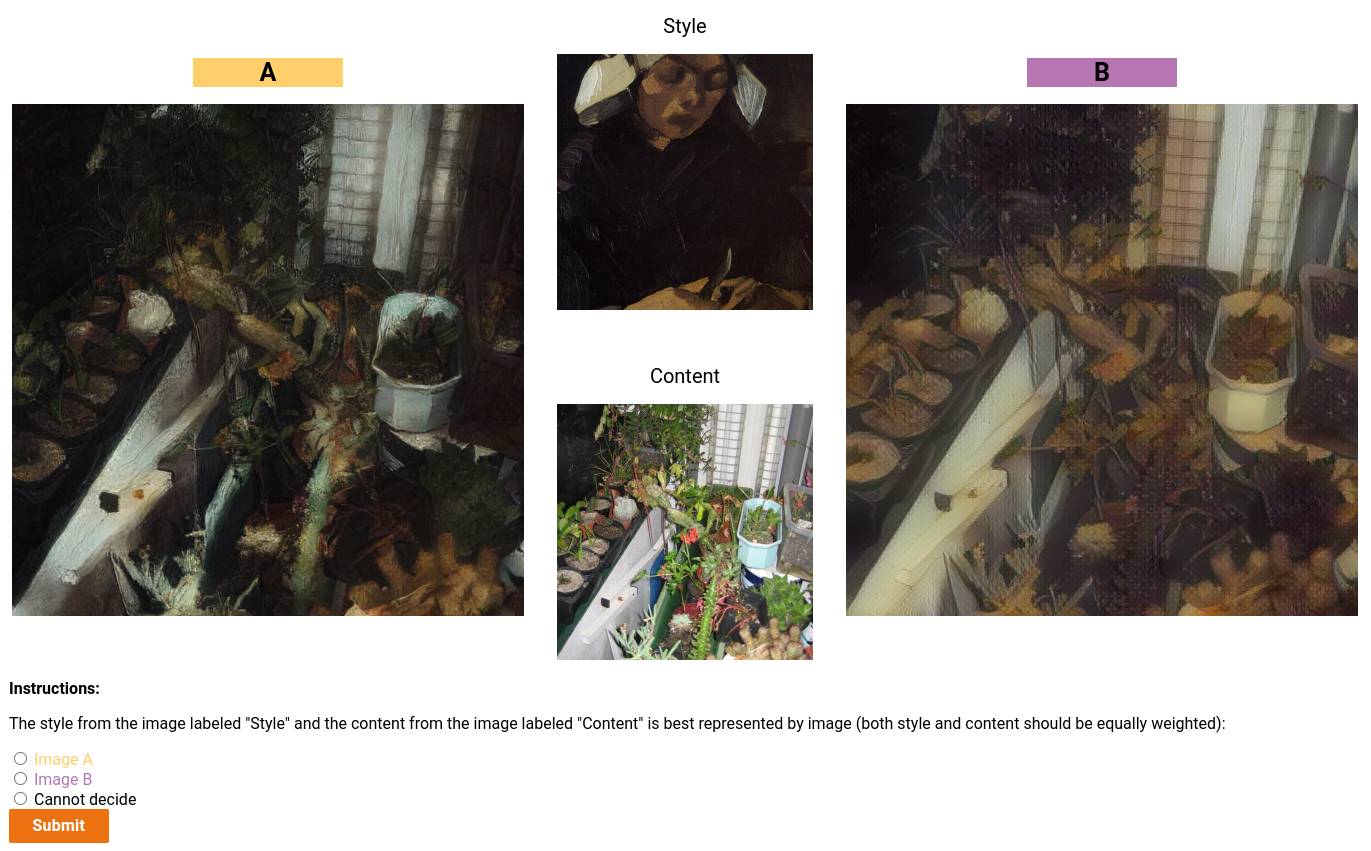}
  \caption{The interface shown to the ATM workers for the user study.}
  \label{fig:user_study}
\end{figure}

In order to test the validity of the ArtFID, we conduct a large-scale human evaluation study on Amazon Mechanical Turk (AMT). We use the same 13 style transfer methods as before. Similar to \cite{Kim2020DST,Kolkin2019}, we chose a pairwise comparison, where the outputs of two methods are compared head-to-head. This choice is in line with recent work from the field of psychology that provides evidence that human perceptual decision-making is significantly impaired when presented with multiple alternatives, rather than just two alternatives \cite{Yeon2015nature}. We compare each possible pairing of style transfer methods. For each pairing, we perform 80 comparisons. Each comparison is voted on by 5 workers and the winner is decided by majority vote. In total there were 31200 tasks and workers were rewarded \$0.03 for each task. Fig.~\ref{fig:user_study} shows the interface that was shown to the AMT workers for a particular task. \\

\subsection{Correlation with Human Judgment}
We evaluate the results of the AMT user study with the widely-used Bradley-Terry model \cite{bradley_terry_model}. Given $m$ entities that are repeatedly compared with each another in pairs, the Bradley-Terry model yields an overall score $\gamma_i$ for each entity $i$ that induces a global ranking of the entities. The model works under the assumption that if we partition the set of entities into two non-empty subsets, some entity in the first subset is compared to some entity in the second subset at least once \cite{hunter_mm_algos}. This requirement is easily satisfied by comparing each entity with all the other entities, as described in Sec.~\ref{sec:amt}.
We can then estimate the scores $\gamma_1,...,\gamma_m$ using the log-likelihood of the Bradley-Terry model

\begin{equation}\label{eq:bradley_terry_likelihood}
    l(\gamma_1,...,\gamma_m) = \sum\limits_{i=1}^m\sum\limits_{j=1}^m [w_{ij} \log(\gamma_i) - w_{ij} \log(\gamma_i + \gamma_j)],
\end{equation}
where $w_{ij}$ denotes the number of times that entity $i$ was preferred over entity $j$, and we assume that $w_{ii} = 0$ and $\sum_i \gamma_i = 1$ \cite{hunter_mm_algos}. There exists a simple iterative algorithm that yields a unique maximum of Eq.~\ref{eq:bradley_terry_likelihood}. See Sec.~\ref{app:btm} for an outline of the algorithm and \cite{hunter_mm_algos} for a study of its convergence properties. 

To measure how well the ArtFID coincides with human judgment, we compute the correlation between the ranking of style transfer methods obtained from the human evaluation study and the ranking of the same methods induced by the ArtFID. A commonly-used measure for this is \textit{Spearman's rank correlation coefficient (Spearman's $\rho$)} \cite{Spearman1904,Spearman1907}, a non-parametric measure of correlation between rankings. Spearman's $\rho$ is calculated by replacing the actual observations with the corresponding ranks in the formula for the correlation coefficient \cite{Kokoska2000}. % TODO: maybe put formula here
Spearman's $\rho$ is often used as a hypothesis test to determine if there is a relation between two random variables \cite{Dodge2008}. For the two-sided test, the null hypothesis is that both random variables are mutually independent and the alternative hypothesis is that there is either a positive or a negative correlation. For the one-sided test, the null hypothesis is that both random variables are mutually independent and the alternative hypothesis is that there is a positive correlation. We report both the Spearman's $\rho$ as well as the $p$-values of the hypothesis tests in Tab.~\ref{tab:rank_correlation}. \\
To validate our results, we rely not only on our AMT user study but also consider user studies from other style transfer works. We observe that the ranking induced by the ArtFID coincides with the majority of user studies reported in the literature \cite{Chen2021ST,Luo2022ConsistentST,Park2019ST,liu2021adaattn,deng2022stytr2,li2018learning,deng2020multiadapt}. For example, Chen et al. \cite{Chen2021ST} report that users preferred IEContraAST \cite{Chen2021ST} over SANet \cite{Park2019ST}, Gatys et al. \cite{Gatys2016}, LST \cite{li2018learning}, AdaAttN \cite{liu2021adaattn}, WCT \cite{Li2017UniversalST}, and Avatar-Net \cite{sheng2018avatar}. Luo et al. \cite{Luo2022ConsistentST} report that users preferred PAMA \cite{Luo2022ConsistentST} over AdaAttN \cite{liu2021adaattn}, MANet \cite{deng2020multiadapt}, SANet \cite{Park2019ST}, MAST \cite{huo2021manifold}, and AdaIN \cite{Huang2017ArbitraryST}. Park and Lee \cite{Park2019ST} report that users preferred SANet \cite{Park2019ST} over Gatys et al. \cite{Gatys2016}, Avatar-Net \cite{sheng2018avatar}, AdaIN \cite{Huang2017ArbitraryST}, and WCT \cite{Li2017UniversalST}, Liu et al. \cite{liu2021adaattn} report that users preferred AdaAttN \cite{liu2021adaattn} over LST \cite{li2018learning}, SANet \cite{Park2019ST}, MAST \cite{huo2021manifold}, AdaAttN \cite{liu2021adaattn}, and Avatar-Net \cite{sheng2018avatar}. Deng et al. \cite{deng2020multiadapt} report that users preferred MANet \cite{deng2020multiadapt} over SANet \cite{Park2019ST} and AdaIN \cite{Huang2017ArbitraryST}.

\subsection{Consistency with Increasing Perturbations}
\begin{figure}
  \centering
  \includegraphics[width=\textwidth]{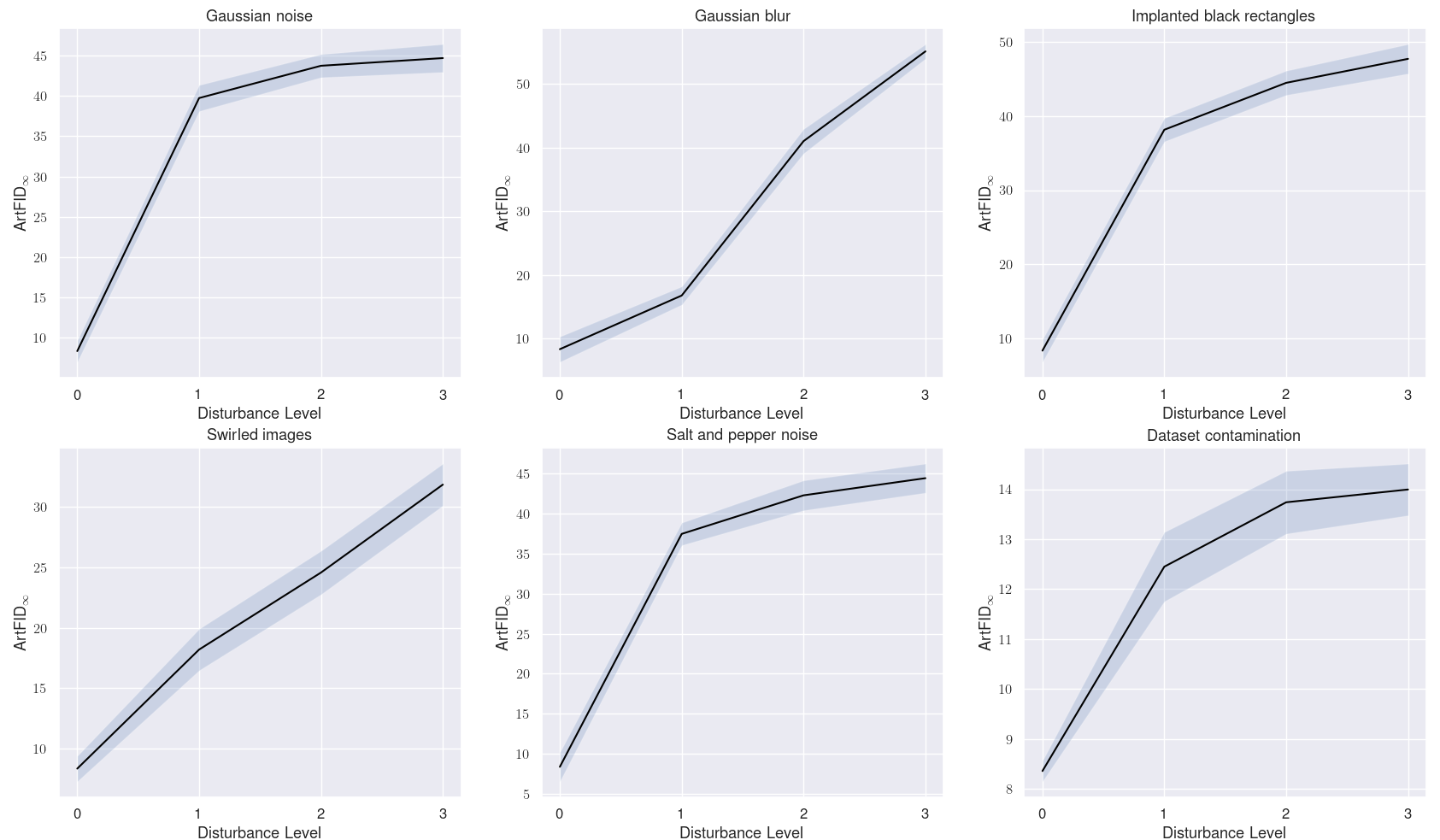}
  \caption{The ArtFID$_\infty$ is consistent with respect to increasing perturbations applied to the stylized images. The stylized images for this experiment were generated using the algorithm proposed by Gatys et al. \cite{Gatys2016}. The line represents the mean and the shaded area the standard deviation over 4 trials.}
  \label{fig:perturbations}
\end{figure}

Heusel et al. \cite{Heusel2017} showed that the FID is consistent with respect to increasing levels of perturbations that are applied to the images. We verify that the ArtFID$_\infty$ inherits this property. Fig.~\ref{fig:perturbations} shows the ArtFID$_\infty$ evaluated using the same image transformations as Heusel et al. \cite{Heusel2017}. The results for the ArtFID are comparable.

\subsection{Ablations}\label{sec:ablation}

We perform the following ablation experiments. All results are reported in Tab.~\ref{tab:ablations}.
\paragraph{Using ImageNet features for ArtFID computation}
Instead of using the modified Inception network trained on the art dataset (Sec.~\ref{sec:data}), we use the ImageNet features from the original Inception network to measure style matching. This shows the effectiveness of learning image representations that reflect stylistic similarity for the style matching task.

\paragraph{Training the Inception network only with artist labels}
We only use the artist labels when training the Inception network on the art dataset. This experiment highlights the importance of harnessing the information from both labels.

\paragraph{Training the Inception network only with style labels}
We only use the style labels when training the Inception network on the art dataset. Again, this shows the importance of using the information from both labels.

\paragraph{Using CLIP for measuring content preservation}
Instead of using the LPIPS metric \cite{zhang2018perceptual} to measure content preservation, we use CLIP \cite{clip2021}. We observe that the LPIPS metric performs slightly better.

\paragraph{Using VGG for measuring content preservation}
Instead of using the LPIPS metric \cite{zhang2018perceptual} to measure content preservation, we use the VGG perceptual loss \cite{Gatys2016,Johnson2016Perceptual}. Again, the LPIPS metric performs slightly better.

% [t]
\begin{table}
  \caption{For each style transfer model, we report the user study score, the ArtFID$_\infty$ and the Deception rate.}
  \label{tab:results}
  \centering
  \begin{tabular}{|l|c|c|c|}
    \hline
    \textbf{Model} & \textbf{User Study Score} $\boldsymbol{\gamma
    }$ $\uparrow$ & \textbf{ArtFID$_\infty$} $\downarrow$ & \textbf{Deception Rate} $\uparrow$ \\
    \hline
    PAMA \cite{Luo2022ConsistentST} & 0.103 & 9.129 $\pm$ 0.478 & 0.307 $\pm$ 0.001 \\
    IEContraAST \cite{Chen2021ST} & 0.105 & 9.495 $\pm$ 0.495 & 0.291 $\pm$ 0.006 \\
    URST \cite{chen2022towards} & 0.092 & 10.846 $\pm$ 0.426 & 0.167 $\pm$ 0.007 \\
    AdaAttN \cite{liu2021adaattn} & 0.093 & 10.911 $\pm$ 0.356 & 0.108 $\pm$ 0.008 \\
    Svoboda et al. \cite{Svoboda2020cvpr} & 0.073 & 11.202 $\pm$ 0.766 & 0.127 $\pm$ 0.008 \\
    MANet \cite{deng2020multiadapt} & 0.08 & 12.095 $\pm$ 0.878 & 0.128 $\pm$ 0.007 \\    
    SANet \cite{Park2019ST} & 0.077 & 12.373 $\pm$ 0.179 & 0.171 $\pm$ 0.003 \\    
    Avatar-Net \cite{sheng2018avatar} & 0.075 & 13.065 $\pm$ 0.844 & 0.089 $\pm$ 0.007 \\
    AdaIN \cite{Huang2017ArbitraryST} & 0.065 & 13.222 $\pm$ 0.549 & 0.096 $\pm$ 0.007 \\ 
    MAST \cite{huo2021manifold} & 0.067 & 14.937 $\pm$ 0.442 & 0.131 $\pm$ 0.005 \\
    LST \cite{li2018learning} & 0.066 & 14.941 $\pm$ 1.062 & 0.146 $\pm$ 0.008 \\
    Gatys et al. \cite{Gatys2016} & 0.054 & 15.707 $\pm$ 0.307 & 0.16 $\pm$ 0.008 \\ 
    WCT \cite{Li2017UniversalST} & 0.05 & 25.495 $\pm$ 0.665 & 0.06 $\pm$ 0.006 \\    
    \hline
  \end{tabular}
\end{table}

% [t]
% \diagbox[width=10em]{\textbf{Metric}}{\textbf{Correlation}}

\begin{table}
  \caption{The correlation between the rankings induced by the different evaluation methods (ArtFID, ArtFID$_\infty$, and Deception rate) and the ranking induced by the AMT user study. The ranking from the ArtFID$_\infty$ strongly correlates with the ranking from the user study.}
  \label{tab:rank_correlation}
  \centering
  \begin{tabular}{|l|c|c|c|}
    \hline
    \textbf{Metric} & \textbf{Spearman's} $\boldsymbol{\rho}$ $\uparrow$ & \textbf{Two-sided $p$-value} $\downarrow$ & \textbf{One-sided $p$-value} $\downarrow$ \\
    \hline
    ArtFID$_\infty$ & \textbf{0.939} & \textbf{1.87e-06} & \textbf{9.39e-07} \\
    ArtFID & 0.934 & 2.99e-06 & 1.49e-06 \\
    Deception Rate & 0.549 & 0.051 & 0.025 \\
    \hline
  \end{tabular}
\end{table}

% [t]
\begin{table}
  \caption{Ablation experiments. (\textsc{a}) ArtFID as described in Sec.~\ref{sec:approach}, (\textsc{b}) the Inception network is trained on ImageNet instead of the art dataset, (\textsc{c}) the modified Inception network is trained using only the artist labels, (\textsc{d}) the modified Inception network is trained using only the style labels, (\textsc{d}) content preservation is measured using CLIP, and (\textsc{d}) content preservation is measured using the VGG perceptual loss.}
  \label{tab:ablations}
  \centering
  \begin{tabular}{|l|c c|c c|c c|}
    \hline
    \multirow{2}{*}{\textbf{Configuration}} &
      \multicolumn{2}{c|}{\textbf{Spearman's} $\boldsymbol{\rho}$ $\uparrow$} &
      \multicolumn{2}{c|}{\textbf{Two-sided $p$-value} $\downarrow$} &
      \multicolumn{2}{c|}{\textbf{One-sided $p$-value} $\downarrow$} \\
    & {\small ArtFID} & {\small ArtFID$_\infty$} & {\small ArtFID} & {\small ArtFID$_\infty$} & {\small ArtFID} & {\small ArtFID$_\infty$} \\
    \hline
    \textsc{a} Full & \textbf{0.934} & \textbf{0.939} & \textbf{2.99e-06} & \textbf{1.87e-06} & \textbf{1.49e-06} & \textbf{9.39e-07}  \\
    \textsc{b} ImageNet features & 0.302 & 0.329 & 0.316 & 0.271 & 0.158 & 0.136 \\
    \textsc{c} Only artist labels & 0.868 & 0.868 & 0.0001 & 0.0001 & 5.95e-05 & 5.95e-05 \\
    \textsc{d} Only style labels & 0.851 & 0.851 & 0.0002 & 0.0002 & 0.0001 & 0.0001 \\
    \textsc{e} CLIP for content & 0.917 & 0.917 & 9.91e-06 & 9.91e-06 & 4.95e-06 & 4.95e-06 \\
    \textsc{f} VGG for content & 0.928 & 0.923 & 4.61e-06 & 6.85e-06 & 2.30e-06 & 3.42e-06 \\
    \hline
  \end{tabular}
\end{table}

\section{Limitations}\label{sec:limitations}
Obtaining accurate ArtFID measurements is computationally expensive. It requires evaluating the Inception network and the LPIPS metric for several thousands of images.

\section{Conclusion}
In this work, we propose a method for quantitatively evaluating style transfer models. The goal of this work is to complement the currently used evaluation schemes by promoting the automatic comparison of different style transfer approaches and to study their respective strengths and weaknesses through in-depth analyses. A given model is evaluated by measuring content preservation and style matching across many different style and content images. Content preservation is quantified using the well-established LPIPS metric. To measure style matching, we first learn suitable image representations using a large-scale art dataset. We then measure the distance between the feature distributions of the stylized images and style images with respect to the first two moments. We evaluate our method by conducting a user study on Amazon Mechanical Turk and show that it strongly coincides with human judgment.

\section{Acknowledgements}
This work has been funded by the Deutsche Forschungsgemeinschaft (DFG, German Research Foundation) within project 421703927.

\FloatBarrier

% Note: if you want to use up all of the allowed space for the paper,
%       the bibliography will start on top of page 13. Furthermore,
%       from page 13 onwards, there will be *only* bibliography, no more
%       figures/tables.
\bibliographystyle{splncs04}
\bibliography{myegbib}

\appendix

\section{Artwork Dataset}\label{app:dataset}
The exact number of images is 241088. There are 109 unique style labels and 595 unique artist labels.

\section{Training}\label{app:training}
For training, the images are resized to 299x299. Commonly used augmentations such as random flipping are applied to the images. We use a weighted cross entropy loss for training, where the weight for a particular class is the reciprocal of the number of examples for that class. This is done to alleviate class imbalances. Tab.~\ref{tab:validation} shows the validation accuracy of the network.

\begin{table}
  \caption{Validation accuracies for the Inception network trained on the art dataset.}
  \label{tab:validation}
  \centering
  \begin{tabular}{|l|c|c|}
    \hline
    \textbf{Classifier Head} & Top-1 Accuracy & Top-5 Accuracy \\
    \hline
    Artist & 0.79 & 0.91 \\
    Style & 0.69 & 0.95 \\
    \hline
  \end{tabular}
\end{table}

\section{Bradley-Terry Model}\label{app:btm}
In this section, we will explain the iterative algorithm that maximizes log-likelihood of the Bradley-Terry model. The explanation is mostly taken from \cite{hunter_mm_algos}. We also refer to \cite{hunter_mm_algos} for a study of the convergence properties.\\
We start with an initial vector {\boldmath$\gamma^{(1)}$}. We use {\boldmath$\gamma^{(1)}$}$ = \{1,...,1\}$ in our experiments. We then iteratively update {\boldmath$\gamma^{(k)}$} as follows:
\begin{equation}
    \gamma_i^{(k + 1)} = W_i \left[ \sum\limits_{j \neq i} \frac{N_{ij}}{\gamma_i^{(k)} + \gamma_j^{(k)}} \right]^{-1},
\end{equation}
where $W_i$ denotes the number of times that entity $i$ was preferred in total and $N_{ij} = w_{ij} + w_{ji}$ is the number of pairings between entity $i$ and entity $j$. If {\boldmath$\gamma^{(k + 1)}$} does not satisfy the constraint $\sum_i \gamma_i^{(k + 1)} = 1$, it is normalized accordingly.

\end{document}